\documentclass[11pt]{article}

\usepackage[final]{acl}

\usepackage{times}
\usepackage{latexsym}
\usepackage{amsmath, amssymb}
\usepackage{float}
\usepackage[T1]{fontenc}

\usepackage[utf8]{inputenc}

\usepackage{microtype}
\usepackage{adjustbox}
\usepackage{inconsolata}
\usepackage{hyperref} 
\usepackage{graphicx}
\usepackage{multirow}
\usepackage{booktabs} 

\title{Thinking-while-speaking: A Controlled, Interleaved
Reasoning Method for Real-Time Speech Generation}

\title{Thinking-while-speaking: A Controlled, Interleaved Reasoning Method for Real-Time Speech Generation}

\author{
  \textbf{Xuan Du}\textsuperscript{1}\thanks{These authors contributed equally to this work.} ,
  \textbf{Qiangyu Yan}\textsuperscript{1}\footnotemark[1] ,
  \textbf{Wenshuo Li}\textsuperscript{1} ,
  \textbf{Borui Jiang}\textsuperscript{1} ,
\\
  \textbf{Changming Xiao}\textsuperscript{1} ,
  \textbf{Han Shu}\textsuperscript{1} ,
  \textbf{Xinghao Chen}\textsuperscript{1}\thanks{Corresponding author.}
\\[4pt]
  \textsuperscript{1} Huawei Technologies
}

\begin{document}
\maketitle
\begin{abstract}
The \textit{thinking-while-speaking} paradigm aims to make AI communication more human. A key challenge is maintaining fluent speech while performing deep reasoning. 
  Our method, \textbf{InterRS}, tackles this by inserting reasoning steps only during natural speech generation. 
  This requires high-quality data where reasoning and speech are precisely aligned, and the length ratio are under controlled.
  We introduce a novel pipeline to generate such seamlessly interleaved audio data. 
  To train our model, we combine interleaved SFT with refined data and reinforcement learning with two new rewards: a \textit{TA-Balance Reward} to manage timing and thinking-answer ratio, and a \textit{Linguistic Quality Reward} to refine expression. 
  Experiments show our approach achieves \textbf{13\%} better performance on mathmatical and logic benchmarks while \textbf{generating instant response} like a spoken-language instruct model which outputs fast CoT response. Furthermore, our method generates more natural and fluent answers than prior methods.
\end{abstract}

\begin{figure*}[ht]
  \begin{center}
     \centerline{\includegraphics[width=\textwidth]{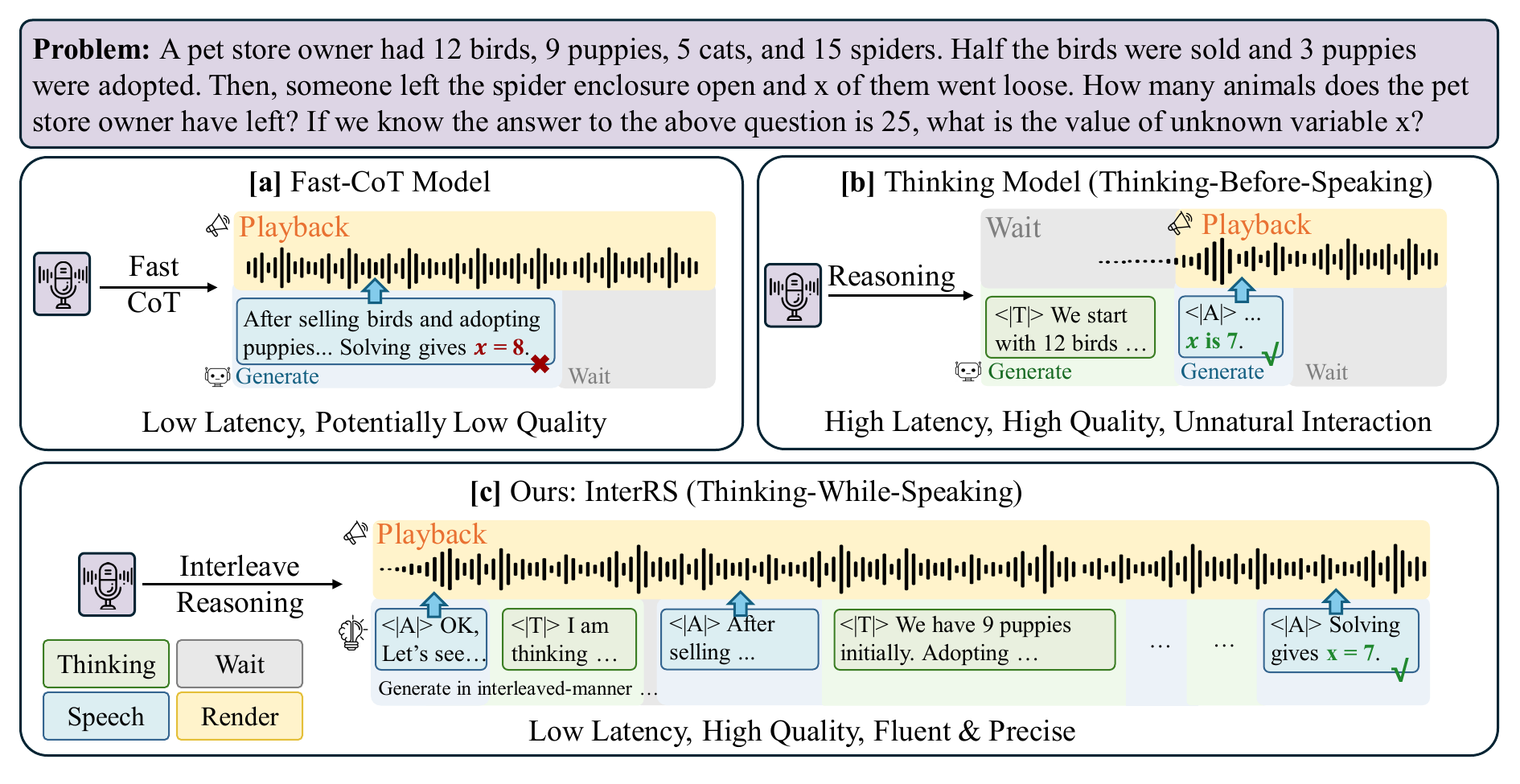}}
  \vspace{-5pt}
    \caption{
      Our framework, InterRS, integrates reasoning steps within temporal gaps in speech generation, enabling seamless thought-speech alignment. 
      It advances beyond traditional interleaved methods by jointly optimizing fluency and accuracy through two novel reward functions, achieving improved performance on reasoning tasks and perceived response naturalness.
    }
    \label{main_fig1}
  \end{center}
  \vspace{-20pt}
\end{figure*}

\section{Introduction}

Spoken language models (SLMs) are rapidly advancing~\cite{lakhotia2021generative, rubenstein2023audiopalm, audiolm2023, speechgpt2023} . 
This progress has made real-time interaction a key research focus, aiming for more natural and human-like communication~\cite{chu2024qwen2audio, gong2024moshi, wang2026unified}. 
When solving complex logical tasks, humans often exhibit a \textit{thinking-while-speaking} pattern~\cite{sacks1974simplest}. 
They continuously perform internal reasoning while expressing their thoughts~\cite{ma2024audiocot}.

As a comparable early attempt to incorporate reasoning capabilities into SLMs, researchers are adapting the chain-of-thought (CoT) paradigm from text models for speech~\cite{wei2022chain,ma2024audiocot,chen2024cotst, kojima2022large}. 
By leveraging reinforcement learning, the model's capacity for reasoning can be further improved.
However, the traditional \textit{thinking-before-speaking} approach faces major challenges in speech interaction~\cite{gong2024moshi}. 
The length of a reasoning chain is often unpredictable. 
Requiring a model to finish all reasoning before speaking causes unacceptably high first token latency. 
This delay breaks the real-time flow of conversation.

To improve the user experience, recent works explore an interleaved \textit{thinking-while-speaking} paradigm~\cite{xie2025miniomnireasoner, yu2024speech_reasoning}. 
The goal is to perform supplementary thinking during speech generation.
Yet, current interleaved methods have two key shortcomings. 
First, the alignment between thinking and speaking content can be rigid.
This makes it hard to balance reasoning depth with verbal fluency. 
Second, training often lacks effective control over the thinking-answer ratio.
This can lead to overly long replies and introduce undesired pause in speech generation.

To solve these problems, we propose a new reasoning paradigm named \textbf{InterRS} for speech models, as show in Figure~\ref{main_fig1}. 
It leverages the temporal gaps in speech generation for supplementary reasoning. 
This allows for significantly more accurate and fluent responses while maintaining real-time performance.
Our main contributions are threefold:
\begin{itemize}
    \item \textbf{A framework for fluent, controllable interleaved reasoning data.}
    We designed a fine-grained data alignment mechanism. 
    It segments and associates speaking content with the underlying thinking process based on semantic logic, allowing control over the thinking ratio of data. 
    It ensures natural transitions and semantic consistency in the interleaved speech.
   
    \item \textbf{A training strategy with two novel rewards.}
    The interleaved SFT phase is designed to endow the model with the foundational capability for controlled output length and fluent articulation.
    For the post-training phase, we introduce two reward functions. 
    The TA-Balance Reward dynamically regulates the interleave reasoning length to prevent over-thinking.
    The Linguistic Quality Reward is optimized to maintain fluent and stable speech flow during reasoning interleaving.
    
    \item \textbf{Strong performance on logical \& mathematical reasoning.}
    Our method shows significant performance gains (+13\% compared to fast CoT responses) and maintaining the capability of instant response on benchmarks. 
    This includes the mathematical benchmarks, such as Spoken-MQA~\cite{DBLP:journals/corr/abs-2505-15000}, and the logical benchmarks, such as SATA-Bench~\cite{DBLP:journals/corr/abs-2506-00643}.
    Crucially, in linguistic quality evaluation, our method generates speech with significantly higher fluency than traditional interleaved reasoning models. 
\end{itemize}

\section{Related Work}

\textbf{Spoken Language Models.} 
The development of speech interaction technology has witnessed a transition from mechanical responses to human-like natural conversation. In early research, speech interaction systems primarily adhered to a cascaded architecture, a paradigm that strictly divides the interaction process into three independent modules: automatic speech recognition (ASR)~\cite{DBLP:journals/inffus/KheddarHH24, DBLP:journals/speech/BenzeghibaMDDEJFLMRRTW07}, natural language model for understanding and generation (LLM)~\cite{DBLP:conf/nips/Ouyang0JAWMZASR22, DBLP:journals/jmlr/RaffelSRLNMZLL20, DBLP:conf/acl/ElhadyAA25, DBLP:conf/nips/KojimaGRMI22}, and text-to-speech (TTS)~\cite{DBLP:journals/corr/abs-2502-06490}. 
However, the linear transmission between modules leads to significant information loss, particularly as paralinguistic information from speech is almost entirely lost during conversion to text. Moreover, the cumulative latency across processing steps makes it difficult to meet the requirements for real-time interaction in terms of time-to-first-token (TTFT)~\cite{liu2025xtalkunderestimatedpotentialmodular}.
With the advent of the large-scale pre-training era, end-to-end speech models have gradually become a central focus of research. These models attempt to process speech signals within a unified semantic space by discretizing audio into speech tokens or leveraging cross-modal alignment techniques, thereby equipping the model with the ability to directly listen and speak~\cite{wang2025generalauditoryintelligencelarge}.

\textbf{Reasoning Paradigms: From TBS to Interleaved Thinking.} To enhance the logical depth of SLMs, researchers introduced the \textit{thinking-before-speaking} (TBS) paradigm, where a model generates a complete hidden CoT before responding. Although accurate, TBS induces prohibitive latency. To bridge this gap, \textit{thinking-while-speaking} or interleaved reasoning has emerged as a more anthropomorphic paradigm. STITCH~\cite{chiang2025stitch} utilizes chunked reasoning by exploiting the computational redundancy during speech synthesis. Mini-Omni-Reasoner~\cite{xie2025miniomnireasoner} proposes a token-level Thinking-in-Speaking formulation using a hierarchical architecture. More recently, Mind-Paced Speaking~\cite{wu2025mindpacedspeaking}  adopts a dual-brain approach, employing a formulation brain for high-level reasoning to pace a separate articulation brain. 

\textbf{Reinforcement Learning for Reasoning.} Reinforcement Learning (RL) has become the dominant paradigm for strengthening LLM reasoning. Prior work demonstrated that models can be guided to interleave thinking and answering through RL, significantly reducing the TTFT~\cite{xie2026interleavedreasoninglargelanguage}. However, granting models the flexibility often leads to reasoning collapse~\cite{tian2025stepaudior1}, where policies discard explicit reasoning for brevity.

\section{Methodology}

We introduce the interleaved reasoning framework, a training paradigm designed to bridge the gap between reasoning accuracy and interaction latency.  
This section outlines our structured data synthesis pipeline, the interleaved supervised fine-tuning strategy, and the subsequent optimization via Group Relative Policy Optimization (GRPO)~\cite{grpo2024, deepseekr12025}. 

\subsection{Interleaved Reasoning}
\textit{Thinking-while-speaking} paradigm enables SLMs to execute reasoning in discrete stages while concurrently delivering verbal summaries. 
We define a complete response sequence $S$ as follows:
\vspace{-5pt}
\begin{equation}
S = {(T_1, A_1), (T_2, A_2), \dots, (T_n, A_n)},
\end{equation}
Where $T_i$ represents the $i$-th thinking segment and $A_i$ represents the corresponding answer segment.
In a real-time deployment scenario, these segments serve distinct operational roles:

\textbf{Background Inference}:  Each $T_i$ represents a discrete stage of internal logical deduction. These segments provide the rigorous thinking foundation necessary to maintain high-fidelity reasoning, significantly improving response accuracy compared to non-reasoning models. 

\textbf{Foreground Playback}: Each $A_i$ serves as a periodical summary that distills the core conclusions of its preceding thinking segment $T_i$. To ensure a natural and fluent user experience, $A_i$ is specifically optimized to be more concise and colloquial than the raw logic in $T_i$, mimicking the brevity and tone of human speech. These segments are streamed to a TTS engine for immediate playback.

By interleaved $A_i$\&$T_i$ segments, 
the generation time for $T_{i+1}$ is masked by the playback time of $A_i$, the system achieves a seamless interaction flow, effectively concealing the computational latency of complex reasoning while preserving the model's logical rigor.

\subsection{Data Construction and Interleaved Refinement}

\begin{figure*}[ht]
  \begin{center}
     \centerline{\includegraphics[width=\textwidth]{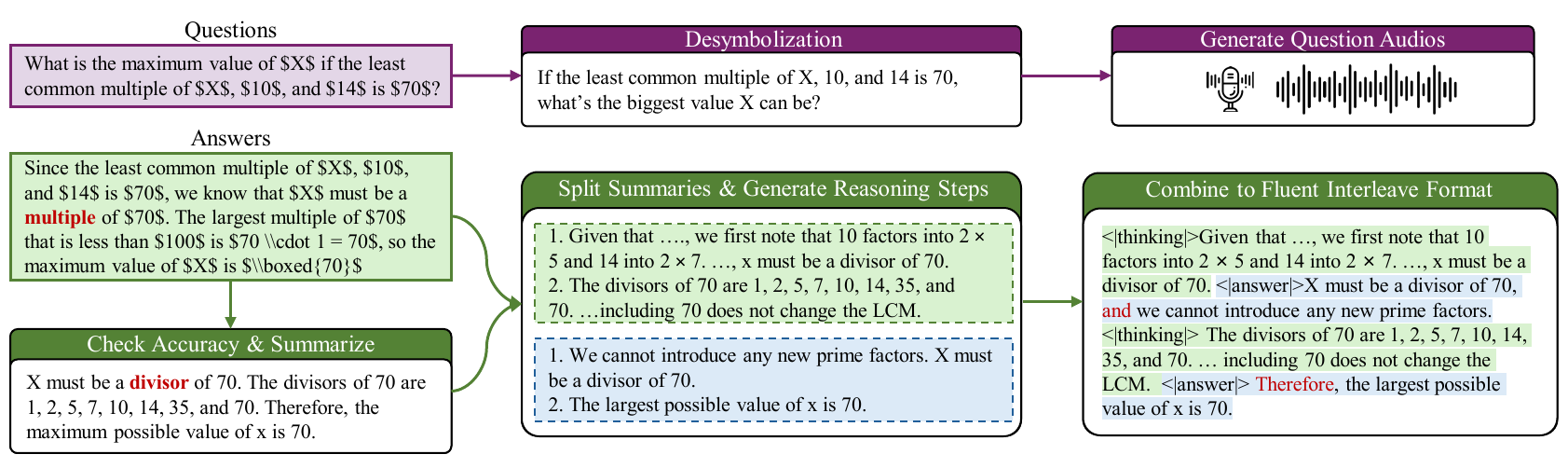}}
\vspace{-5pt}
    \caption{
      Data Pipeline: (The top line) For questions, the original question first undergoes desymbolization and colloquial normalization to improve its suitability for spoken delivery, followed by speech synthesis via TTS. (The bottom line) For answers, an initial accuracy validation is performed, after which the content is condensed into a summary. This summary is then segmented into semantically coherent units. These segments are subsequently utilized to retrieve corresponding reasoning steps paired with their original answer segments. Finally, the matched thinking-answer pairs are synthesized into an interleaved reasoning response, while simultaneous linguistic refinement ensures the overall coherence and quality of the final output.
    }
    \label{main_fig2_data}
  \end{center}
\vspace{-25pt}
\end{figure*}

To construct a dataset that supports both high-quality reasoning and fluent speech, we developed a three-stage pipeline to restructure and align raw data into the interleaved reasoning format as shown in Figure~\ref{main_fig2_data}. 

\textbf{Logical Verification and Oral-Style Summarization}
\underline{Answer Part}: We verify the logical integrity of reasoning paths and reconstruct flawed trajectories. These reasoning paths are then distilled into concise, oral-style summaries that preserve essential logic while adopting a natural spoken tone and removing formal writing artifacts.
\underline{Question Part}: Symbolic questions undergo desymbolization into conversational queries to enhance auditory clarity. We then employ CosyVoice2 \cite{DBLP:journals/corr/abs-2412-10117} to convert these refined questions into natural-sounding audio inputs.

\textbf{Semantic Unit Splitting}
To create natural thinking windows within the voice stream, we split the summary into several small, independent speech units $\{A_1, A_2, \dots, A_n\}$ based on logical completeness and natural pauses in speech. Each unit is designed to represent a single sub-point or a short sentence that a person would typically say in one breath. This segmentation allows us to insert thinking segments between spoken fragments while maintaining a smooth logical progression.

\textbf{Controlled-Ratio Thinking Content Construction}
For each speech unit $A_i$, we synthesize a corresponding internal thinking segment $T_i$ based on the original logic. To ensure the computational latency is effectively masked during on-device deployment, we enforce a length-ratio constraint.
The ratio needs to ensure that under standard on-device inference speeds, the generation of the next reasoning block $T_{i+1}$ is fully covered by the audio playback duration of the current segment $A_i$. 
The final output is a structured sequence of $(T_i, A_i)$ pairs that emulate a natural \textit{thinking-while-speaking} cognitive flow.
Here we use the ratio of 4:1 for thinking:answer.

\subsection{Interleaved SFT}
Given the structural complexity of frequent logical transitions in the interleaved format, applying RL directly to a base model would result in extremely sparse reward signals and unstable training trajectories. To mitigate this, we implement Supervised Fine-Tuning (SFT) before proceeding to RL optimization. The primary objective of this phase is to align the model’s output distribution with the structural requirements of the interleaved reasoning format. 

Conventional thinking implementations typically rely on paired delimiters (e.g. \texttt{<think>...</think>}) to encapsulate reasoning. In contrast, our framework utilizes singular state-transition flags to minimize sequence overhead. We define two specialized tokens: the trigger \texttt{<|thinking|>} signals the model to transition from verbalization to internal logical deduction, and the trigger \texttt{<|answer|>} signals a switch from reasoning back to generating a conversational response. This design treats each token as a functional operator that marks the boundary of a state change within the interleaved topology. By using single-token flags instead of paired tags, we effectively compress the overall sequence length and improve inference efficiency—saving valuable tokens in long-form multi-turn dialogues. 

The SFT stage serves as a crucial initialization step. By training on the high-quality synthetic data described in the previous section, the model acquires foundational instruction-following capabilities while internalizing the \textit{thinking-answer} alternating pattern. This ensures that when the model enters the RL phase, it already possesses a high degree of format compliance. Establishing this stable baseline prevents the model from collapsing during early RL iterations and significantly enhances the convergence efficiency of the policy gradient optimization.

\subsection{Reinforcement Learning}

To further optimize the quality 
of outputs, this paper employs the GRPO algorithm. This algorithm enables policy updates without requiring an extra Critic network by sampling a group of candidate outputs and computing relative rewards.

\subsubsection{Reward Function Modeling}
The overall reward function \(R_{total}\) is designed as a weighted sum of a TA-Balance Reward, an Accuracy Reward, and a Linguistic Quality Reward:
\vspace{-5pt}
\begin{equation}
    R_{total} = \omega_{TA} \cdot R_{TA} + \omega_{Acc} \cdot R_{Acc} + \omega_{LQ} \cdot R_{LQ},
\end{equation}
where \(\omega\) represents the hyperparameter weights for each reward component.

\textbf{TA-Balance Reward  (\(R_{TA}\))}
The model must strictly adhere to the interleaved reasoning topology. Any violation of the format (e.g., consecutive thinking blocks or missing tags) results in a zero reward. Additionally, to prevent over-thinking on simple tasks and ensure the thinking segments fit within the audio playback window, we introduce a quadratic length penalty. For the \(i\)-th thinking segment, its format score \(r_i\) is calculated as follows:
\vspace{-5pt}
\begin{equation}
    r_{i} = \max \left( 0, 1 - \left( \frac{L_i - L_{target}}{L_{target} / 2} \right)^2 \right),
\end{equation}
where \(L_i\) is the word count of the current thinking segment, and \(L_{target}\) is the preset word count threshold. The final format reward is the arithmetic mean of all thinking segment scores.

The quadratic decay provides a smoother gradient signal. This soft constraint encourages the model to stay near the $L_{target}$ without being overly penalized for minor deviations, ensuring a stable optimization process.

\textbf{Accuracy Reward (\(R_{Acc}\))}
The accuracy reward serves as a hard constraint on logical correctness. we extracts the final prediction \(y_{pred}\) from the content following the last \texttt{<|answer|>} token and compares it with the ground truth \(y_{gt}\):
\vspace{-5pt}
\begin{equation}
\scalebox{0.96}{$
    R_{Acc} = \mathbb{I}(y_{pred} = y_{gt}) =
    \begin{cases}
    1, & \text{if } y_{pred} = y_{gt} \\
    0, & \text{otherwise}
    \end{cases}
    $}
\end{equation}

\textbf{Linguistic Quality Reward  (\(R_{LQ}\))}
 
The interleaved reasoning format inherently risks fragmenting the verbal output, potentially leading to semantic disjointedness or lack of narrative flow between separated speech segments. To counteract this, we aim to enforce that the final concatenated response, denoted as:
\vspace{-5pt}

\begin{equation}
\mathcal{A} = A_1 \oplus A_2 \oplus \cdots \oplus A_n,
\end{equation}
maintains the coherence of a natural-language sentence. We leverage the linguistic priors of a reference model to guide this process~\cite{grpocare}. The core premise is that a fluent and semantically consistent response should exhibit low perplexity (and thus high likelihood) under the distribution of a frozen reference model. For a given question $\mathcal{Q}$ and a concatenated full answer $\mathcal{A}$, we compute its log-likelihood under the reference distribution $p_{\text{ref}}$:
\vspace{-5pt}
\begin{equation}
L(\mathcal{A}) =  \sum_{t=1}^{|\mathcal{A}|} \log p_{\text{ref}}(a_t \mid \mathcal{Q}, a_{<t}),
\end{equation}
A higher value of $L(\mathcal{A})$ indicates that the generated sequence aligns closely with natural language patterns learned by the reference model. For a candidate $k$ within a group $\mathcal{G}$, we compute its normalized likelihood score $\tilde{L}^{k}$ and the Linguistic Quality Reward is formulated as:
\begin{equation}
\scalebox{0.85}{$
R_{LQ}^{k} = \mathbb{I}(k \in \mathcal{C}) \cdot \max\left( 0, \beta \cdot \left( \tilde{L}^{k} - \text{mean}_\mathcal{G}(\tilde{L}) \right) \right),
$}
\end{equation}
where $\mathcal{C}$ represents the subset of samples that arrived at the correct final answer, and $\beta > 0$ is a scaling factor. We apply this reward exclusively to correct samples to prevent the model from optimizing for fluency at the expense of logical accuracy. Additionally, a KL penalty is omitted to avoid overly constraining the model’s exploratory capability, thereby enabling the integration of this reward at virtually no additional cost relative to the standard GRPO framework. By explicitly rewarding high-likelihood sequences, we encourage the model to bridge the gaps between interleaved segments, ensuring that the final output constitutes a coherent and logically unified narrative suitable for high-quality speech synthesis.

\section{Experiments}
\subsection{Implementation Details}
We employ Qwen2.5-Omni-3B~\cite{xu2025qwen25omnitechnicalreport} as the baseline model to leverage its inherent multimodal capabilities. The fine-tuning process focuses on the thinker module to cultivate interleaved reasoning paths for complex logical and mathematical tasks. The training set comprises approximately 12,000 high-quality samples from K\&K~\cite{DBLP:journals/corr/abs-2410-23123} and MetaMath~\cite{yu2024metamathbootstrapmathematicalquestions}, which are pre-processed into the interleaved reasoning format to ensure a balanced ratio between internal thought segments and audible responses. The training pipeline proceeds in two sequential phases. First, a interleaved SFT phase initializes the interleaved output structure and establishes foundational reasoning habits. This stage runs for 3 epochs with a learning rate of $4 \times 10^{-6}$ under a cosine decay schedule. It is followed by a RL phase lasting 2 epochs, using a group size of $G=16$ and a learning rate of $1 \times 10^{-6}$ . 

\subsection{Evaluation Metrics}

To comprehensively evaluate performance in spoken interaction scenarios, we focus on three core dimensions: Accuracy, Fluency, and Temporal Efficiency. 

\subsubsection{Accuracy and Generalization}
For in-domain evaluation, we use the test sets of K\&K and MetaMath. To examine out-of-distribution generalization, we employ Spoken Multi-Question Answering (SMQA)~\cite{DBLP:journals/corr/abs-2505-15000} and the SATA-Bench~\cite{DBLP:journals/corr/abs-2506-00643}. SMQA evaluates mathematical reasoning across four distinct categories, including Short Digit (S) and Long Digit (L) for arithmetic operations, alongside Single-step Reasoning ($R_1$) and Multi-step Reasoning ($R_m$) for mathematical logic tasks.  Given the varying difficulty and sample sizes, we calculate a weighted aggregate score to represent the overall performance: 
\vspace{-5pt}
\begin{equation}
Score_{total} = \sum_{i \in {S, L, R_1, R_m}} \frac{N_i}{N_{total}} \cdot Score_i,
\end{equation}
where $N_i$ represents the number of questions in each branch. We also use SATA-Bench, which covers domains such as law, medicine, and logic reasoning. In this benchmark, the model must correctly identify all valid options in a multiple-choice setting. This benchmark is particularly rigorous because human accuracy is only 17.93\%, providing a strong baseline for evaluating advanced reasoning capabilities.

\subsubsection{Fluency Assessment}
To ensure the interleaved format does not compromise the quality of the audible response, we concatenate the answer segments $\{A_1, A_2, \dots, A_n\}$ into a single text. We then utilize DeepSeek-V3 as an automated judge to score the overall fluency on a scale of 0 to 2:

\textbf{0 (Incoherent):} Logical contradictions exist between segments. For example, a model might first state ``The total loss is 48 units" and later claim ``Therefore, the net loss is 32 units" without reconciliation.

\textbf{1 (Fluent but Disjointed):} Logical flow is correct, but linguistic transitions between segments are abrupt. For example, the reply ``She spends \$10 on orange creamsicles. She spends \$4.50 on ice cream sandwiches. She spends \$14.50 in total." might have a correct solving process, but there is no smooth transition between sentence.

\textbf{2 (Excellent):} Both logic and expression are seamless and coherent.

\subsubsection{Temporal Efficiency}
Immediate responsiveness is critical for the user experience in real-time conversational AI. In our evaluation, we employ Instant Response as a categorical metric to denote whether a model facilitates immediate streaming interaction or necessitates high-latency pre-computation.

\subsection{Main Result}

We conducted a comprehensive evaluation of the proposed \textit{interleaved reasoning} paradigm and various baseline models. The performance across logical and mathematical reasoning benchmarks is detailed in Table~\ref{tab:main_results}.

\begin{table*}[t]
  \centering
  
  \setlength{\tabcolsep}{3pt}
  \begin{tabular}{llcccccc}
    \hline
    Type & Method
    & \multicolumn{2}{c}{Logic} & \multicolumn{2}{c}{Mathematical} & Avg. $\uparrow$ & Inst. Resp. \\
    \cmidrule(lr){3-4} \cmidrule(lr){5-6}
    & & SATA-Bench & KK-audio & SMQA & Meta-audio & & \\
    \hline
    \multirow{2}{*}{Thinking} 
    & Zero-shot & 0.30 & 5.57 & 70.11 & 52.60 & 32.15 & $\times$ \\
    & SFT+RL & \underline{24.00} & \underline{42.14} & \textbf{76.74} & \textbf{61.50} & \textbf{51.10} & $\times$ \\
    \hline
    Direct & Zero-shot & 0.20 & 6.57 & 42.15 & 10.50 & 14.86 & $\surd$ \\
    \hline
    \multirow{2}{*}{Fast CoT} 
    & Zero-shot & 0.10 & 6.57 & 69.40 & 19.70 & 23.94 & $\surd$ \\
    & SFT+RL & \textbf{25.20} & 28.57 & 53.19 & 40.70 & 36.92 & $\surd$ \\
    \hline
    \multirow{3}{*}{Interleaved} 
    & Mini\_OR & - & - & 69.01 & - & - & $\surd$ \\
    & SFT & 17.50 & 39.14 & 73.01 & 59.10 & 47.19 & $\surd$ \\
    & \textbf{InterRS} & 21.10 & \textbf{42.57} & \underline{74.43} & \underline{59.20} & \underline{49.33} & $\surd$ \\
    \hline
  \end{tabular}%
   \vspace{-5pt}
  \caption{\label{tab:main_results} \textbf{Main Result on Logical and Mathematical Benchmarks.} We report accuracy scores ($\uparrow$) across four datasets. The \textbf{bolded result} indicates the highest score, and the \underline{underlined result} indicates the second highest score. Results demonstrate that our InterRS achieves competitive accuracy while maintaining the capability of instant response, bridging the gap between reasoning and real-time interaction.}
  \vspace{-10pt}
\end{table*}
  
\subsubsection{Models and Baselines }
To provide a robust comparison, we categorize the compared methods into four distinct paradigms based on their reasoning and response patterns:

\textbf{Thinking}: This paradigm follows the traditional \textit{thinking-before-speaking} approach. \textbf{Zero-shot} utilizes the vanilla Qwen2.5-Omni-3B, prompted to generate a full reasoning chain before producing the answer. \textbf{SFT+RL} is further fine-tuned and optimized with reinforcement learning to improve the performance. 

\textbf{Direct}: This paradigm constrains the model to output only the final answer without any explicit reasoning steps. 

\textbf{Fast CoT}: This paradigm aims for providing a concise, \textit{fast-thinking} summary. The \textbf{Zero-shot} variant is elicited through specific prompting, whereas \textbf{SFT+RL} is specifically trained to optimize this fast-thinking behavior. 

\textbf{Interleaved}: This paradigm integrates reasoning and speech into a single, unified stream. \textbf{Mini\_OR}~\cite{xie2025miniomnireasoner} is an external state-of-the-art baseline for interleaved reasoning; \textbf{SFT} is trained with supervised fine-tuning to master the interleaved structure; \textbf{InterRS} is our final model, optimized from the SFT base using our proposed method.

\subsubsection{Analysis of Key Findings}
The result suggests that our proposed InterRS model effectively reconciles reasoning rigor with real-time interaction. It achieves a highly competitive average score of 49.33, closely approaching the accuracy of monolithic Thinking SFT+RL model (51.10). Crucially, InterRS attains this while maintaining instant response capability. In contrast, the Thinking baseline, despite its marginal accuracy lead, requires full pre-computation, rendering it unsuitable for smooth conversation. This demonstrates that our InterRS preserves over \textbf{96.5\%} of the analytical capability of Thinking SFT+RL model without incurring its latency costs.

 Furthermore, the comparison highlights a trade-off: models  of Thinking, Direct and Fast CoT that prioritizing speed suffer significant accuracy drops on complex tasks. Conversely, the Thinking models that achieve high accuracy sacrifice all real-time responsiveness. InterRS uniquely demonstrates how high reasoning accuracy and low latency can be jointly optimized for conversational AI.

\section{Ablation Studies }
In this part we train and test the models only on mathematical datasets.

\subsection{Analysis of Length and Interleaving Controls} 
\begin{figure}[ht]       
  \centering
  \includegraphics[width=\linewidth]{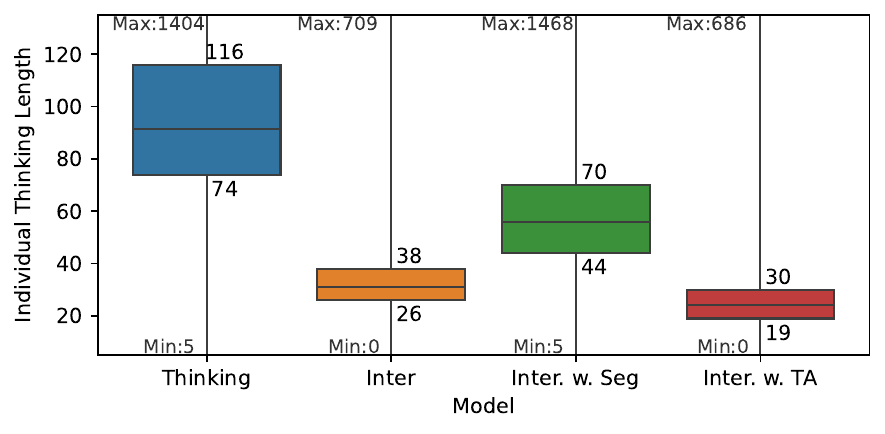}
\vspace{-22pt}
  \caption{Distribution of \textbf{individual thinking segment lengths} across different model configurations.}
\vspace{-5pt}
  \label{fig:thinking_box}
\end{figure}

We evaluated the impact of our TA-Balanced rewards against three alternative configurations to understand how different optimization objectives influence reasoning behavior. The \textbf{Thinking} model follow the \textit{thinking-before-speaking} format. The \textbf{Inter} model enforces correct tag structures but permits unconstrained thinking segment lengths. The \textbf{Inter. w. Seg} rewards the model for reducing the total number of interleaved transitions under the correct interleaved format, which may alleviate over thinking on simple questions. Our \textbf{Inter. w. TA} encourages the model to generate thinking segment under target length and keep a balanced length ratio between thinking segments and answer segments.
We examine the stability of individual thinking segments. Figure~\ref{fig:thinking_box} reveals that the Thinking model is unsuitable for real-time interaction due to excessive latency, evidenced by an IQR of 42 tokens. The Inter. w. Seg and Inter baseline models lack rhythmic consistency, as they consolidate reasoning into obstructive blocks with IQRs of 26 and 12 tokens, respectively.
\textbf{The Inter. w. TA model achieves the most refined control}, yielding the narrowest IQR of only 11 tokens and the lowest median. This statistical stability effectively eliminates the risk of long-duration reasoning blocks, providing the predictable and fluent interaction cadence required for a controlled real-time framework.


\begin{table}[t]
  \centering
  \resizebox{\columnwidth}{!}{
  
  \vspace{-8pt}
  \begin{tabular}{lccc|cc}
    \hline
    Model & Fmt & Seg & TA & Acc & Len \\
    \hline
    Thinking & $\surd$ & × & × & \textbf{69.40} & 120.87 \\
    Inter & $\surd$ & × & × & 68.63 & 128.64 \\
    Inter w.Seg & $\surd$ & $\surd$ & × & 69.02 & 123.61 \\
    Inter w.TA & $\surd$ & × & $\surd$ & 68.65 & \textbf{104.88} \\
    \hline
  \end{tabular}
  }
    \vspace{-5pt}
\caption{\label{tab:ablation_len}Ablation studies across total thinking segment lengths and interleaving times controls.}
  \vspace{-5pt}
\end{table}

\begin{table}[t]
  \centering
  
  \vspace{-5pt}
        \begin{tabular}{lcccc}
    \toprule  
 No. & Smooth & LQ Rew &   Acc & Fluency \\    
 \midrule    
(1) & $\times$ & $\times$ &  68.65  & 1.65  \\
(2) & $\surd$ & $\times$ &   68.32  & 1.74  \\
(3) & $\surd$ & $\surd$ &   \textbf{69.29}  & \textbf{1.83}  \\
    \bottomrule
    \end{tabular}
      \vspace{-5pt}
\caption{  \label{tab:ablation_smooth}
Ablation study of \textbf{conversational smoothing and Linguistic Quality reward}.}
  \vspace{-10pt}
\end{table}

\begin{figure*}[ht]
  \begin{center}
     \centerline{\includegraphics[width=\textwidth]{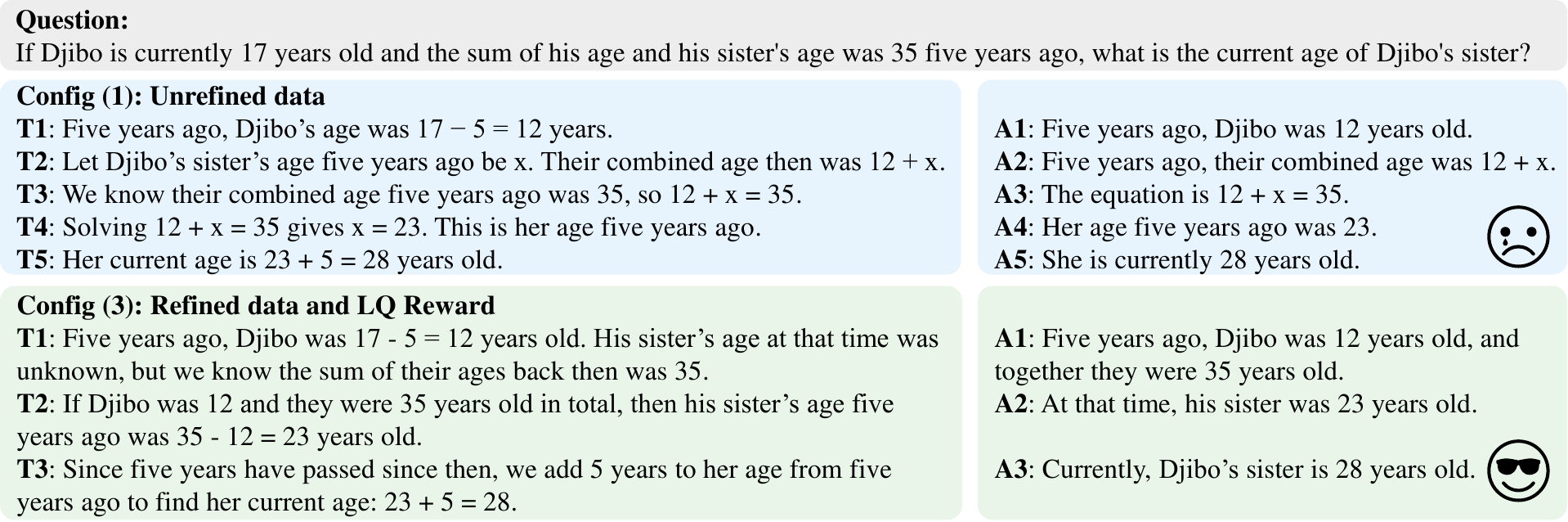}}
     \vspace{-5pt}
    \caption{
      Comparison of responses from Configuration (1) and Configuration (3) for a multi-step reasoning task.
    }
    \label{fig_case}
  \end{center}
  \vspace{-20pt}
\end{figure*}

\begin{figure}[t]

  \begin{center}
    \vspace{-5pt}
     \centerline{\includegraphics[width=\columnwidth]{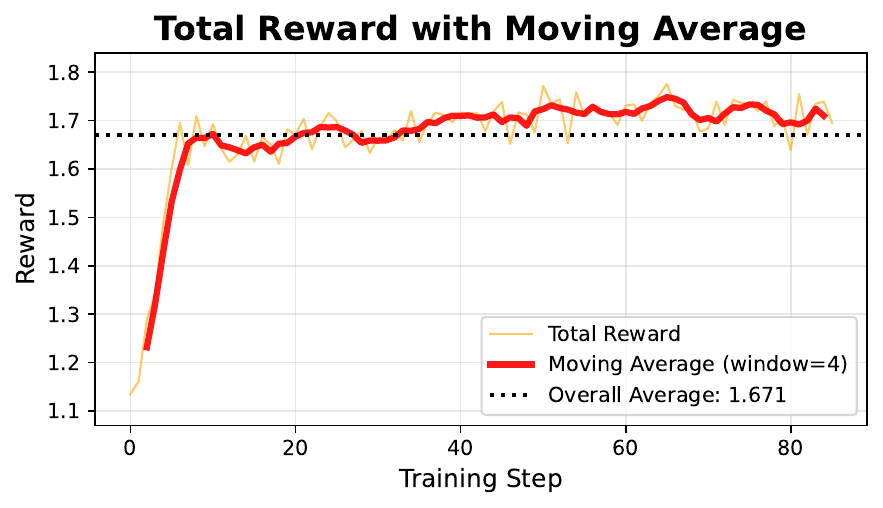}}
    \vspace{-12pt}
    \caption{
      Rewards of our InterRS in training process.
    }    
    \label{fig_reward}
  \end{center}
\vspace{-25pt}
\end{figure}

According to Table~\ref{tab:ablation_len}, while the Inter. w. Seg configuration yields a slightly higher accuracy, it results in a significantly higher total thinking token count. In contrast, our Inter. w. TA model maintains competitive performance while achieving the shortest total thinking length. 

This reduction in total thinking length significantly enhances the system's inference efficiency. A lower total token count translates to higher reasoning density, where the model learns to identify the most direct logical path while minimizing redundant computation.

\subsection{Impact of Data Refinement and LQ Reward} 
We further investigate the contributions of our data refinement pipeline and the Linguistic Quality Reward, with all configurations built upon the TA-Balanced Reward foundation. Configuration (1) serves as the baseline, corresponding to the Inter. w. TA model from the preceding analysis.


According to Table~\ref{tab:ablation_smooth}, the transition from Configuration (1) to (2) demonstrates that training on our refined data significantly enhances the naturalness of the spoken output, increasing the fluency score from 1.65 to 1.74. The qualitative shift toward more human-like verbal expressions confirms the effectiveness of the data refinement pipeline. Building upon the smoothed data, \textbf{InterRS} (Configuration 3) introduces the Linguistic Quality Reward, which \textbf{leads to the best overall performance}, achieving a peak fluency score of 1.83 and an improved accuracy of 69.29. This progression confirms that explicitly rewarding the semantic alignment between internal reasoning and audible segments is essential. This ensures that the fragmented response constitutes a coherent, logically unified answer rather than a series of disjointed statements. The reward curve is shown on Figure~\ref{fig_reward}.

Figure~\ref{fig_case} illustrates the qualitative advantages of InterRS through a case study. Configuration (1) produces fragmented, isolated segments that lack natural flow and introduces an undefined variable $x$ directly into the audible output. Conversely, Configuration (3) achieves superior semantic continuity by synthesizing reasoning steps into cohesive, fluent sentences. By employing transitional phrases (e.g., "At that time") to anchor logical inferences, it ensures mathematical accuracy while optimizing the delivery for human-like speech communication.

\section{Conclusion}
In this work, we present a novel training method designed to achieve instant response while maintaining high-quality thinking in spoken language models. By using the delay in speech generation to mask thinking tokens, our approach eliminates the prohibitive latency typical of conventional \textit{thinking-before-speaking} models.

Our approach preserves high analytical accuracy while ensuring instantaneous response capability. We achieve fine-grained control over thinking lengths through the TA-Balanced reward. By eliminating the risk of unexpectedly long reasoning blocks, this mechanism ensures a stable interaction cadence. Furthermore, the synergy between conversational data refinement and the Linguistic Quality reward allows the model to achieve SOTA fluency and accuracy. These optimizations ensure that the interleaved response stream constitutes a coherent, logically unified answer rather than a series of disjointed statements, validating the feasibility of integrating high-level reasoning with temporally efficient speech generation.
 \newpage
\section{Limitations}
While our model achieves instant response times resembling a spoken-language instruct model, the current framework evaluates the interaction in a primarily static turn-taking format. Real-world human communication is highly dynamic, often involving interruptions or abrupt topic shifts. How the interleaved reasoning segments adapt or truncate when the external speech flow is interrupted by the user is an area that requires dedicated architectural extensions in future work.

\bibliography{custom}

\appendix

\end{document}